\documentclass[10pt,twocolumn,letterpaper]{article}

\usepackage[pagenumbers]{cvpr} % To force page numbers, e.g. for an arXiv version

\usepackage{graphicx}
\usepackage{amsmath}
\usepackage{amssymb}
\usepackage{booktabs}
\usepackage{bm}

\usepackage[dvipsnames]{xcolor}
\usepackage{comment}
\usepackage{multirow}

\usepackage{enumitem}
\setlist{nosep}

\newcommand{\websitelink}{https://stanford-tml.github.io/circle_dataset/}
\newcommand{\website}{\href{\websitelink}{website}}
\makeatletter
\renewcommand\paragraph{\@startsection{paragraph}{4}{\z@}%
                                    {1mm}%
                                    {-1em}%
                                    {\normalfont\normalsize\bfseries}}
\makeatother

\usepackage[pagebackref,breaklinks,colorlinks]{hyperref}

\usepackage[capitalize]{cleveref}
\crefname{section}{Sec.}{Secs.}
\Crefname{section}{Section}{Sections}
\Crefname{table}{Table}{Tables}
\crefname{table}{Tab.}{Tabs.}

\def\cvprPaperID{7029} % *** Enter the CVPR Paper ID here
\def\confName{CVPR}
\def\confYear{2023}

\begin{document}

\title{CIRCLE: Capture In Rich Contextual Environments}

\begin{comment}
\author{João Pedro Araújo \\
Stanford University\\
\and
Jiaman Li\\
Stanford University\\
\and
Karthik Vetrivel\\
Stanford University\\
\and
Rishi Agarwal\\
Stanford University\\
\and
Jiajun Wu\\
Stanford University\\
\and
Deepak Gopinath\\
Meta AI\\
\and
Alexander Clegg\\
Meta AI\\
\and
C. Karen  Liu\\
Stanford University\\
}
\end{comment}

\author{João Pedro Araújo$^1$, Jiaman Li$^1$, Karthik Vetrivel$^1$, Rishi Agarwal$^1$, Jiajun Wu$^1$,\\
Deepak Gopinath$^2$, Alexander Clegg$^2$, C. Karen  Liu$^1$ \\
$^1$Stanford University, $^2$Meta AI\\
}

% Trick from https://tex.stackexchange.com/questions/55764/input-a-figure-between-title-and-body-in-twocolumn-form
\twocolumn[{%
\renewcommand\twocolumn[1][]{#1}%
\maketitle
\begin{center}
    \centering\vspace{-12pt}
    \captionsetup{type=figure}
    \includegraphics[width=\linewidth]{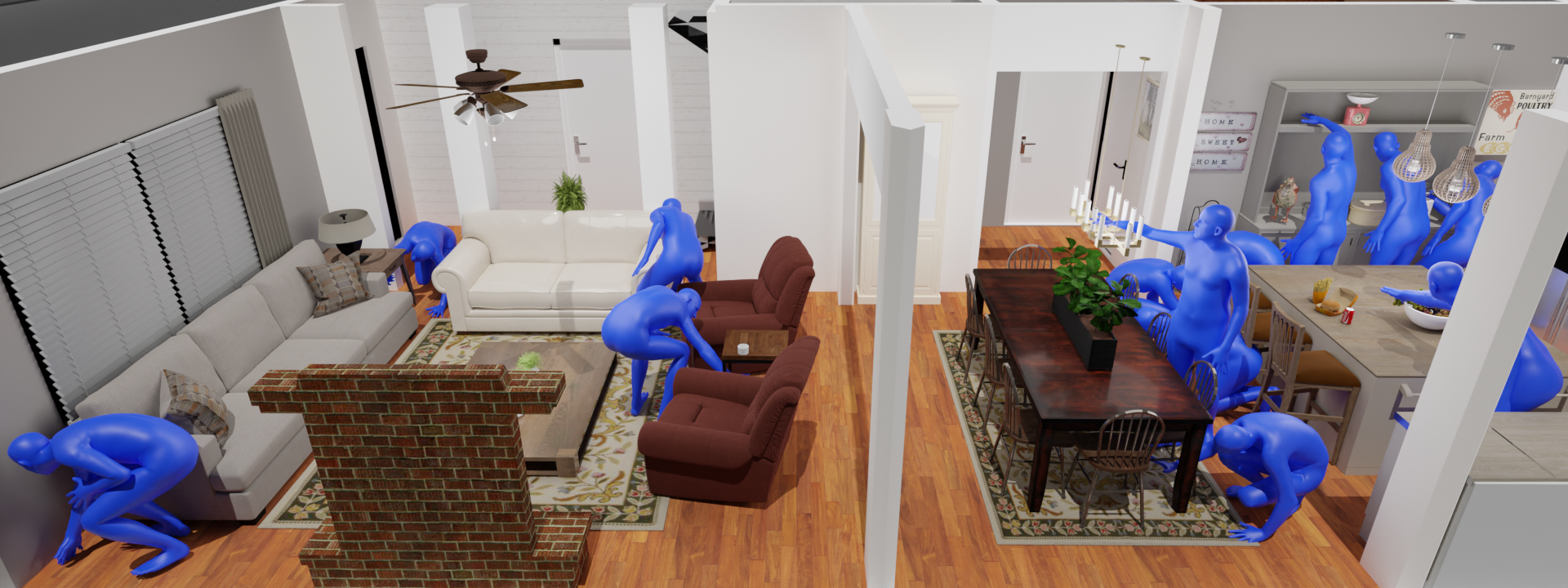}
    \captionof{figure}{Example poses from CIRCLE captured from real human motion in a virtual environment.}
    \label{fig:teaser}
\end{center}%
}]

%%%%%%%%% ABSTRACT
\begin{abstract}
\vspace{-7pt}
Synthesizing 3D human motion in a contextual, ecological environment is important for simulating realistic activities people perform in the real world. However, conventional optics-based motion capture systems are not suited for simultaneously capturing human movements and complex scenes. The lack of rich contextual 3D human motion datasets presents a roadblock to creating high-quality generative human motion models. We propose a novel motion acquisition system in which the actor perceives and operates in a highly contextual virtual world while being motion captured in the real world. Our system enables rapid collection of high-quality human motion in highly diverse scenes, without the concern of occlusion or the need for physical scene construction in the real world. We present CIRCLE, a dataset containing $10$ hours of full-body reaching motion from $5$ subjects across nine scenes, paired with ego-centric information of the environment represented in various forms, such as RGBD videos. We use this dataset to train a model that generates human motion conditioned on scene information. Leveraging our dataset, the model learns to use ego-centric scene information to achieve nontrivial reaching tasks in the context of complex 3D scenes. To download the data please visit our \website.

\end{abstract}

\section{Introduction}
\label{sec:intro}

Humans excel at interacting with complex environments, effortlessly engaging in everyday tasks such as getting out of a car while carrying a backpack or plugging a power cord into an outlet behind a cabinet. The remarkably flexible and compliant human body enables access to narrow or cluttered spaces where clear paths are not available. Synthesizing 3D human motion that reflects this ability to navigate in highly contextual, ecological environments, such as our homes, grocery stores, or hospital operating rooms, will significantly impact applications in Embodied AI, Computer Animation, Robotics, and AR/VR.   

Machine learning models have significantly advanced the creation of 3D human motion and behaviors in recent years. However, the success of the ML-approach hinges on one condition---the human motion data for training models must be of high quality, volume, and diversity. Traditional motion capture (mocap) techniques focus on the ``human movement'' itself, rather than the state of the environment in which the motion takes place. While mocap can faithfully record human kinematics, capturing humans in a contextual scene requires physical construction of a production set and specific props in the capture studio. This steep requirement limits the capability of today's mocap technologies to holistically capture realistic human activities in the real world. 

We propose to eliminate the costly requirement of physical staging by capturing human motion during interactions with a \emph{virtual reality simulation}. This allows us to capture motion like the ones shown in Figure \ref{fig:teaser}, where a person reaches into cluttered spaces in a furnished apartment. Additionally, we are able to simultaneously record paired first-person perspectives of the virtual environment through VR, as illustrated in Figure \ref{fig:real_habitat_mosh_comparison}. With paired ego-centric observation of the world, we can now train motion models to not only comprehend the \emph{how} of certain tasks, but also the \emph{why} behind an individual's movements. 

By creating the complex scene in the virtual world and keeping the capture space in the real world empty, our method provides four crucial advantages over state-of-the-art solutions. First, creating a highly contextual environment in VR is much simpler and less costly than in actual reality. Second, capturing the state of the real world requires complex sensor instrumentation, while the state of the virtual world is readily available from the simulator. Third, because the capture space in reality is always empty, our system is not subject to occlusions that degrade the motion quality, regardless of any clutter in the perceived environment. Fourth, the data acquired by such a system provide 3D human motions and corresponding videos of the environment rendered in any camera view of choice, such as the egocentric view.

We use a Meta Quest 2 headset and the AI Habitat simulator in our experiments. However, our system is agnostic to the choice of hardware, simulator, and virtual environment. To illustrate the possibilities enabled by the availability of contextual motion capture data, we collect a dataset, CIRCLE, containing ten hours of full-body reaching motion within nine indoor household scenes. CIRCLE contains challenging reaching scenarios, including reaching for an object behind the toilet, between tightly placed furniture, and underneath the table. Finally, we use CIRCLE to train a model that generates reaching motions conditioned on scene information. Our model takes as input the starting pose of the person in the scene as well as the target location of the right hand, and automatically generates a scene-aware sequence of human motion that reaches the target location. We propose two different methods to encode the scene information and compare them against baselines. 

In summary, the contributions of this work include:
\begin{itemize}
    \item A novel motion acquisition system to collect 3D human motion with synchronized scene information,
    \item A novel dataset, CIRCLE, with $10$ hours of human motion data from $5$ subjects in $9$ realistic apartment scenes,
    \item A data-driven model, trained on CIRCLE, for generating full-body reaching motion within an environment.
\end{itemize}

\section{Related Work}
\label{sec:related}

\paragraph{Human Motion Datasets.} The desire to thoroughly model human motion has contributed to many high-quality datasets. High-resolution optical motion capture datasets \cite{de2009guide, sigal2010humaneva, loper2014mosh,  Trumble2017TotalCapture, AMASS:2019} range from the smaller CMU Motion Capture Database \cite{de2009guide} to AMASS \cite{AMASS:2019}, a rich human motion collection that unifies several mocap datasets and contains over 40 hours of motion data. Other works have modeled a vast diversity of motions, including whole-body reaching, object manipulation \cite{GRAB:2020}, human-scene contact \cite{hassan2019prox, huang2022}, human-chair interaction \cite{zhang2022couch}. While these datasets provide an excellent baseline for analyzing human motion and interaction with specific objects, they do not consider the constraints of the 3D environments humans naturally move through, a key contribution of our approach.

Early work from Hasler~\etal~\cite{Hasler:2009} recovers joint configurations with scene constraints using multiple unsynchronized moving cameras. More recently, GPA \cite{wang2019geometric} and SAMP ~\cite{SAMP:2021} capture scene-conditioned human motion by placing a limited set of physical objects in the mocap area. Our work obviates the need for physical construction of environments by placing the actor in a virtual world, enabling capture in diverse and cluttered scenes. BEHAVE ~\cite{bhatnagar2022behave} provides multi-view RGBD videos with 3D pose and contact information to enable tracking of human-object interactions. GIMO and EgoBody ~\cite{zheng2022gimo, zhang2021egobody} propose motion datasets with IMU-based and egocentric image capture respectively. Moreover, both datasets offer 3D reconstructions of the scenes. However, IMU-based motions can result in drifting ~\cite{jiang2022transformer} and LiDAR and camera-based 3D reconstructions are inherently noisier than the virtual ground truth.

\paragraph{3D Human Motion Synthesis.}
In recent years, significant progress has been made in synthesizing kinematic human motion. Early progress saw the introduction of periodic motion embeddings \cite{ormoneit2001, urtasun2006, sidenbladh2000} and non-linear autoregressive architectures for pose tracking and modeling \cite{arikan2002, pavlovic2001, wang2005, taylor2007}. Recurrent neural networks \cite{shu2019, wang2018, martinez2017, fragkiadaki2015, kim2017} and similar autoregressive models have seen success on smaller sets of motion. Conditional variational autoencoders (cVAE) \cite{rempe2021humor, ling2020character, cai2021} are another popular architecture for generating plausible motion sequences. Lately, Transformers \cite{aksan2021, song2022, petrovich2021, seong2021} and diffusion models \cite{tevet2022human, zhang2022motiondiffuse} have seen the most success at generating unseen motions. Inspired by recent progress, we train a Transformer-based model architecture on our dataset to generate scene-conditioned motion.  

\paragraph{Constrained Pose Generation.}
Our task of synthesizing 3D human motion in a contextual, ecological environment is inherently a constrained full-body pose generation challenge. Much of the work in constrained pose generation involves 3D hand-object manipulation \cite{Kim:2014:SHS, zhang2021manip}. Although not as rigorously explored, constrained full-body pose generation is not novel. Gupta~\etal~\cite{gupta2008pose} propose an observation-driven Gaussian process latent variable model for human pose estimation using information from scenes and actions. Cao~\etal~\cite{cao2020} tackle full-body scene-aware motion synthesis by addressing the global trajectory and local pose prediction. The authors of the previously mentioned SAMP dataset \cite{SAMP:2021} also propose and train an algorithm to generate interactions with chairs, sofas, and tables. Another paper, COUCH \cite{zhang2022couch}, builds on the SAMP method and provides users with fine-grained control over the generation of human-chair interactions. Recent work includes using Transformer architectures for gaze-directed and context-aware human motion \cite{zheng2022gimo} and estimating 3D body-scene contact from a single image \cite{huang2022}. Another recent related work is GOAL \cite{taheri2021goal}, a data-driven procedure to generate the full-body grasping motion conditioned on an object's geometry and location relative to the body.

\section{Data Collection}
\label{sec:data}

\begin{figure}
    \centering
    \includegraphics[width=\columnwidth]{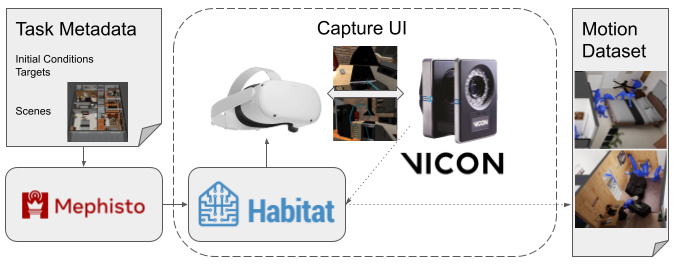}
    \caption{An overview of the data collection system, including the Mephisto experimentation framework, a VR headset (Meta Quest 2 in our case), AI Habitat simulator, and Vicon motion capture.}
    \label{fig:data_system_overview}
\end{figure}

\begin{figure}
    \centering
    \includegraphics[width=\columnwidth]{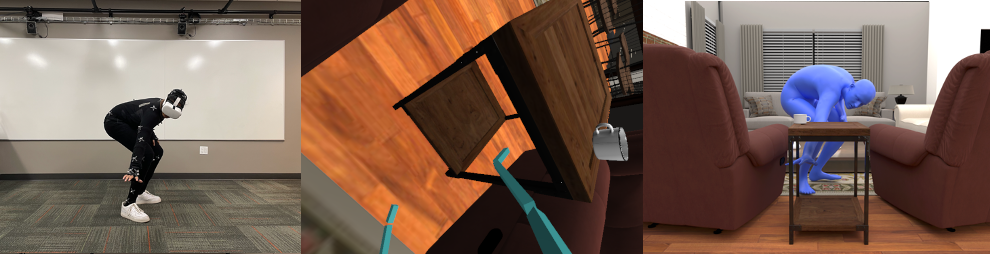}
    \caption{The motion capture process both in the real world (left) and the Habitat VR app (middle). Right: the corresponding SMPL-X mesh generated by MoSh and rendered in Blender.}
    \label{fig:real_habitat_mosh_comparison}
\end{figure}

\begin{figure}
    \centering
    \includegraphics[width=\columnwidth]{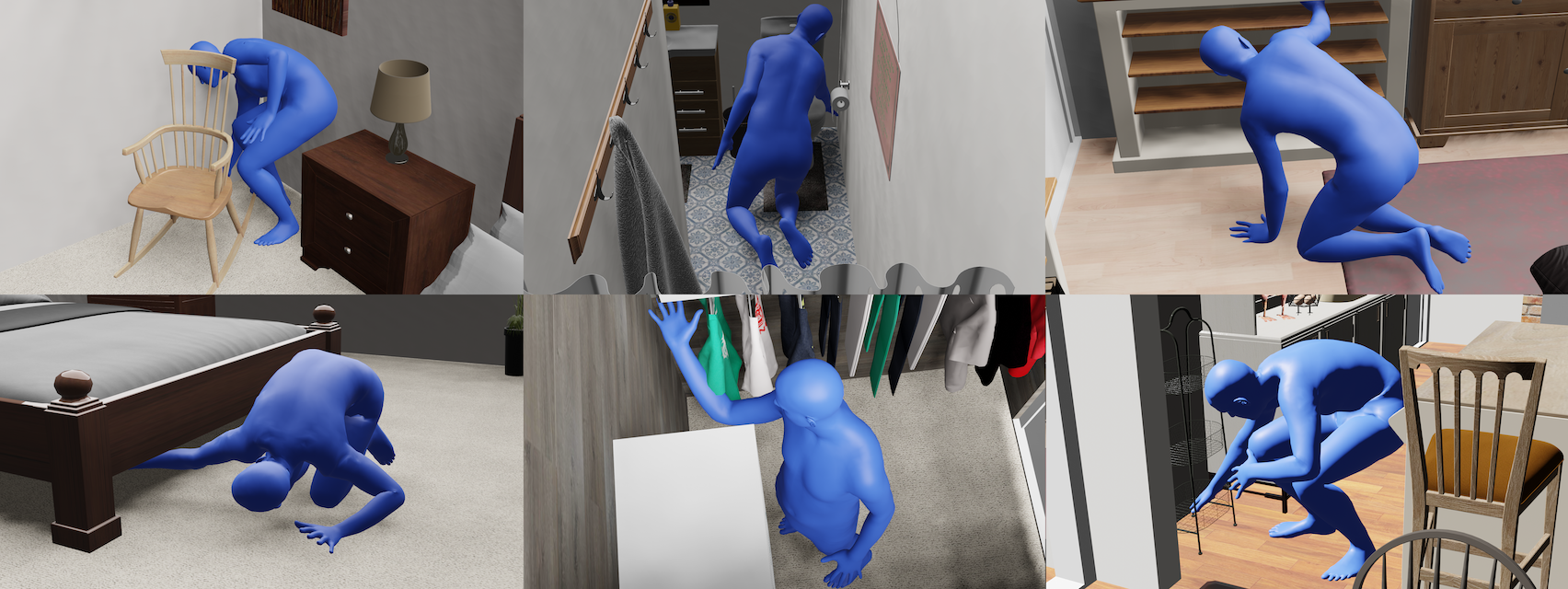}
    \caption{Examples of diverse poses found in CIRCLE.}
    \label{fig:breadth_of_poses}
\end{figure}

\begin{figure}
    \centering
    \includegraphics[width=0.4\columnwidth]{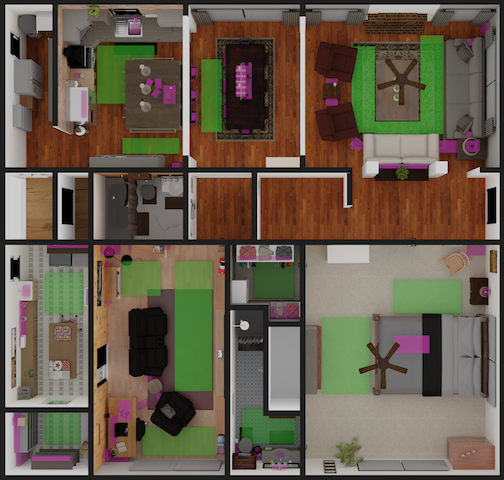}
    \hfill
    \includegraphics[width=0.58\columnwidth]{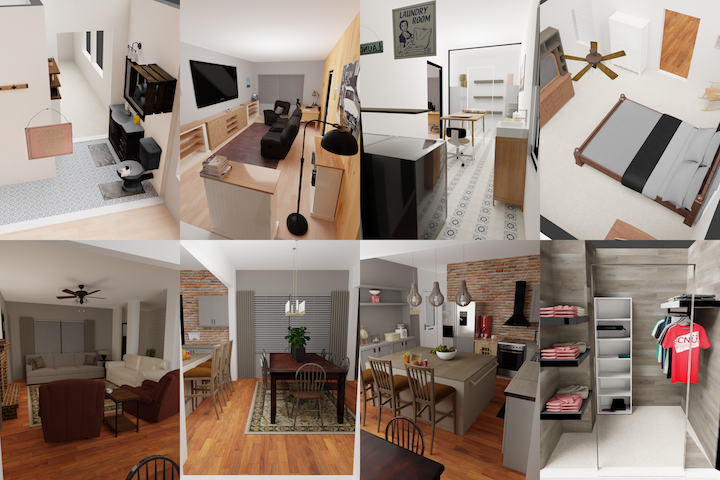}
    \hfill
    \caption{Left: Layout of the initial state (green) and goal (purple) sampling regions within the scenes. Right: Representative images of the following scenes (left to right, top to bottom): bathroom, media room, laundry, bedroom, living room, dining room, kitchen, and closet.}
    \label{fig:start_goal_annotations}
\end{figure}

Our motion acquisition system is designed to enable collection of a contextual motion capture dataset with high volume, high diversity, and high quality.

\paragraph{High Volume.} Using a VR headset to collect a large-scale human dataset poses a unique challenge due to the concerns of user discomfort and costly context switching. The VR app must run as fast as possible to minimize motion sickness and ensure a smooth user experience. In addition, having a highly streamlined pipeline to capture motion in large batches is crucial for minimizing overheads and scaling up the dataset. Thus, an ideal system should maximize the efficiency of data collection during the active VR time. 

\paragraph{High Quality.} Although the data collected using optics-based motion capture are often of high quality, our system depends on the virtual world being immersive and realistic such that it does not influence the natural behaviors of the actor. Being untethered and hands-free is absolutely necessary. In addition, we need to give the actor visual feedback, such as the sight of their virtual body, whenever possible to ensure the naturalness of their performance. 

\paragraph{High Diversity.} Our system should support various subjects, scenes, and tasks. We need tools that assist the planning and preparation process for capturing diverse motion sequences. We also need a streamlined pipeline to work with a variety of virtual worlds and assets.

\subsection{System Overview} 
\label{sec:data:system}

With the previous design goals in mind, we built a system to collect contextual motion capture data (\cref{fig:data_system_overview}). Our system has two main components: the motion capture system and the VR app which provides the actor with the illusion of being within a virtual scene. We use AI Habitat, an open-source embodied AI research platform \cite{habitat19iccv,szot2021habitat}, as our virtual world simulator. Habitat provides easy access to a wide variety of interaction-ready virtual environments and assets, as well as an API tailored to Embodied AI experimentation. We build a WebXR app on top of Habitat's web build \cite{rramrakhya2022}, capable of running at real-time rates on a VR headset with a modern web browser installed. This VR app streamlines all data collection logic, allowing the user to focus on quickly executing capture tasks. Data recorded by the app (headset trajectories and simulation state) are submitted to Mephisto \cite{mephisto} for storage.

For capturing human motion, we use a Vicon system with $12$ cameras controlled with Vicon Shogun. The cameras record at $120$ FPS. We connect Shogun and the VR app through a webserver running locally on the machine capture machine. The VR headset interfaces with this webserver through the local WiFi network. This allows the app to directly control the beginning and ending of individual mocap recordings. Conversely, we use Vicon's DataStream SDK to stream the reconstructed skeleton with very low latency to the VR app, allowing users of the system to see their skeleton in the virtual world. This provides visual feedback for events such as collisions, and enables the user to interact more immersively with the virtual world.

\Cref{fig:real_habitat_mosh_comparison} shows an actor using our system to collect a sequence. The room has no obstacles, so all the actor's movements (left image) are in response to what is being shown in the headset (middle image). The image on the right shows what the same sequence looks like after post-processing. We describe in detail each component of our system in the following sections.

\subsection{Preparation} 
\label{sec:data:preparation}

We define \emph{sequences} as individual \emph{clips} of captured data (the atomic unit of CIRCLE). Sequence specifications are generated from pairs of manually annotated start and goal regions (see \cref{fig:start_goal_annotations} for an example), from which we sample concrete instances of start and goal position pairs. We refer to each start/goal region pair as a \emph{task}. Specifically, any two sequences with start and goal positions sampled from the same pair belong to the same task.

We frontload all task generation steps to the preparation phase at the start of each collection session. The goal of this phase is to prepare a list of sequences (individual clips) to be collected. To streamline the specification of these sequences, we develop a tool that allows users to load scenes into Blender and annotate both start and goal regions. We then sample pairs of points from these regions to generate sequence specifications. Sampled goal positions are rejected if they are unreachable (either too far from navigable surfaces or inside scene or object geometry). Finally, we sort the generated sequences to minimize the transition overhead of resetting the scene and importing assets.

\subsection{Collection} 
\label{sec:data:collection}

Given a list of sequences, we simultaneously start the VR app and a Mephisto process. When entering VR for the first time each day, the user must calibrate the headset in order to align it to the Vicon skeleton. Since we do not know the true offset between the headset frame of reference and the captured head bone, we use an alignment heuristic. For details, please see the Supplementary Material. 

After calibration, the user can start recording. Clips containing the sequences listed for collection are labeled as reaching clips. The transition between sequences can also contain valid contextual mocap data. Therefore, instead of discarding them, the system records and labels them separately. As the user completes sequences, the data are logged to the Mephisto server. When the user has finished all pre-generated sequences, the VR app exits and Mephisto closes, caching all sequences resulting from the session.

\subsection{Post-processing}
\label{sec:data:processing}

After collection, CIRCLE data are passed through a variety of post-processing steps.

\paragraph{Mocap data processing.} We use Shogun Post to process and export the captured clips to BVH and C3D formats.

\paragraph{Offline synchronization} Due to latency between the headset and the webserver communicating with the Vicon machine, the start times of the headset and mocap data are misaligned and must be synchronized. By assuming that the offset between the head bone and the headset remains constant during each sequence, this can be accomplished by solving for the time offset which maximizes the convolution between the velocity profiles of the head bone and headset. With start times aligned, we then trim the sequences to the same length and linearly interpolate the headset poses such that every mocap frame has a corresponding headset pose.

\paragraph{Human mesh fitting.} We run MoSh++ \cite{AMASS:2019} (henceforth referred to as MoSh) on the C3D files to acquire the SMPL-X parameters corresponding to each frame in the sequence (\cref{fig:real_habitat_mosh_comparison}, right).

\paragraph{Synthetic sensor information.} After synchronization, we load both the mocap data in BVH format and the VR trajectories in Habitat and extract synthetic sensor information such as ego-centric RGB-D videos (\cref{fig:real_habitat_mosh_comparison}, middle). Additionally, we can use Blender to render first-person RGB-D videos with the SMPL-X meshes calculated by MoSh.

\paragraph{Quality assurance.} Identifying and fixing sequences with artifacts is a demanding manual process. We find that our pipeline has a very high yield of data that does not need to be fixed, so our focus is on identifying sequences with problems so that they can be collected again. To help prioritize, we develop a suite of tools that automatically check for common problems, such as:

\begin{itemize}
    \item \textbf{Task completion.} The task in the VR app is considered complete if the wrist of the live-streamed skeleton is within $1$ cm of the goal location. We use this to validate the accuracy of our pipeline. All collected clips pass this test.
    \item \textbf{Marker swaps}. For a given marker, given its position at frame $t$, we find the nearest marker at frame $t+1$. If the labels of those two markers differ, we consider that those two markers are swapped. Clips that are flagged by this procedure are automatically discarded.
    \item \textbf{Jumps in joint values.} To help spot sequences with jumps in joint values, for each sequence we record the maximum joint linear acceleration in global space and the maximum angular velocity in local space. Sequences that score high values on either of these metrics are then manually inspected.
\end{itemize}

\subsection{Dataset}
\label{sec:data:dataset}

\begin{table}[t]
    \centering
    \setlength{\tabcolsep}{5.5pt}
\begin{tabular}{lcccccc}
    \toprule
    \multirow{2}{*}{Scene} & \multicolumn{3}{c}{Frames} & \multicolumn{3}{c}{Minutes} \\
    \cmidrule(lr){2-4}\cmidrule(lr){5-7}
     & T & R & Total & T & R & Total \\ 
    \midrule
    Bathroom        & 137k & 188k & 325k & 19 & 26 & 45 \\ 
    Bedroom         & 203k & 309k & 512k & 28 & 43 & 71 \\ 
    Clothes closet  & 137k & 210k & 347k & 19 & 29 & 48 \\ 
    Dining room     & 213k & 309k & 522k & 30 & 43 & 72 \\ 
    Kitchen         & 261k & 309k & 570k & 36 & 43 & 79 \\ 
    Laundry room    & 137k & 201k & 338k & 19 & 28 & 47 \\ 
    Laundry closet  & 128k & 216k & 344k & 18 & 30 & 48 \\ 
    Living room     & 165k & 279k & 444k & 23 & 39 & 62 \\ 
    Media room      & 376k & 529k & 904k & 52 & 73 & 126 \\ 
    \midrule
    Total &  &  & 4306k &  &  & 598 \\ \hline
\end{tabular}
    \caption{Breakdown of our dataset by scene and type of clip (\textbf{T}ransition and \textbf{R}eaching). We collect over $4$ million frames, and $10$ hours of data.}
    \label{tab:data_per_scene}
\end{table}

We use our data acquisition system to produce CIRCLE, a dataset of whole body reaching motion within a fully furnished virtual home. Each distinct room in the apartment is considered a separate scene. We manually annotate the start and goal regions for each scene ($128$ tasks in total; see \cref{fig:start_goal_annotations}) using the tool described in \cref{sec:data:preparation} and follow the procedure outlined in \cref{sec:data:collection} for data collection. Physics simulation was disabled during the collection of CIRCLE, which means that the environment is static in all sequences.

\paragraph{Dataset contents.}CIRCLE contains $10$ hours (more than $7000$ sequences) of both right and left hand reaching data for five subjects across $9$ scenes (\cref{fig:start_goal_annotations}, right). The breakdown of data per scene is listed in \Cref{tab:data_per_scene}. The diversity of scenes induces a wide range of motions (\cref{fig:breadth_of_poses}), including reaching high places, bending, crawling, crouching, kneeling, and lying down.

After post-processing, each sequence in CIRCLE includes:
\begin{itemize}
    \item The SMPL-X parameters of a body model fit to the mocap data using MoSh,
    \item The VR headset trajectory synchronized to the mocap skeleton,
    \item Egocentric RGB-D video (rendered with both Habitat and Blender),
    \item Task specific data, such as initial and goal conditions, and the scene where the data are collected.
\end{itemize}

\section{Motion Generation}
\label{sec:motion_generation}

\begin{figure*}[t!]
\begin{center}
\vspace{-5mm}
\includegraphics[width=\textwidth]{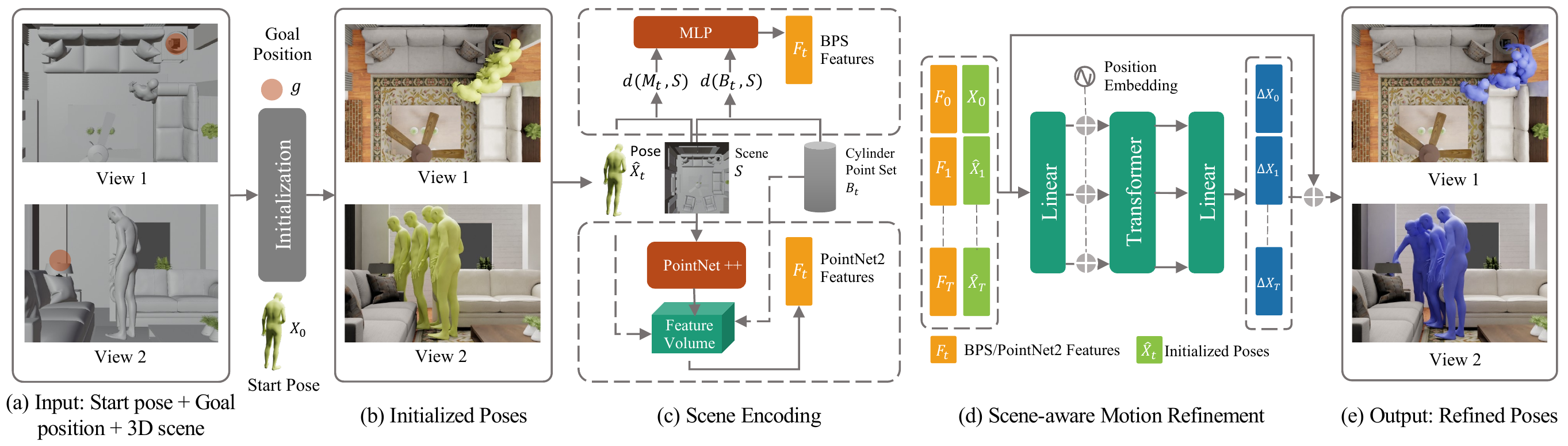}
\end{center}
\vspace{-6mm}
\caption{Method overview. Given a start pose, goal position, and 3D scene (a), we first initialize the input poses using constant local joint rotation and linearly interpolated root translation (b). We extract scene features for each time step using BPS or PointNet++ from the initialized poses, a fixed point set sampled from a cylinder, and scene point clouds (c). We then concatenate scene features and initialized poses and feed them to a transformer-based model (d) to generate final poses (e).}
\label{fig:overview}
\vspace{-1mm}
\end{figure*}

Our goal is to generate a sequence of scene-aware human poses $\bm{X} \in \mathbb{R}^{T \times D}$ given a start pose $\bm{X}_0$, target wrist joint position $\bm{g} \in \mathbb{R}^{3}$ and a scene point cloud $\bm{S} \in \mathbb{R}^{N \times 3}$, where $T$ represents the sequence length, $D$ represents the dimension of the pose state and $N$ represents the number of scene points. The pose state at time step $t$, $\bm{X}_t$, consists of root translation $\bm{p}_t \in \mathbb{R}^{3}$, global joint positions $\bm{J}_t \in \mathbb{R}^{22 \times 3}$, and local joint rotation $\bm{R}_t \in \mathbb{R}^{22 \times 6}$ represented using continuous 6D vectors~\cite{zhou2019continuity}. We use the SMPL-X model~\cite{pavlakos2019expressive} with 22 joints. Our approach is to learn a motion refinement model that takes as input a rough initial motion sequence and scene features, and outputs a scene-aware and higher-quality motion sequence (Figure \ref{fig:overview}).

\paragraph{Motion Initialization.} Given the full body pose at the first time step, $\bm{X}_0$, we generate a naive initial motion sequence $\bm{X}_0, \bm{\hat{X}}_1, \bm{\hat{X}}_2, ..., \bm{\hat{X}}_T$, using a simple procedure. We define a constant pose $\bm{X}_c$ shared by the whole dataset. We first translate $\bm{X}_c$ such that the human's wrist reaches $\bm{g}$ and make this translated pose $\bm{\hat{X}}_T$.We then linearly interpolate the root translation from $\bm{X}_0$ to $\bm{\hat{X}}_T$ to create the in-between frames,  $\bm{\hat{X}}_t$, where $1\leq t \leq T-1$. 

\paragraph{Scene Encoding.} One crucial design decision for training our model is the scene representation. The high-quality privileged 3D information provided by CIRCLE makes many choices of scene encoding possible. We explore two types of scene encoding: Basis Point Set (BPS)~\cite{prokudin2019efficient} representation and features extracted by PointNet++~\cite{qi2017pointnet++}. Both types of scene encoding consist of human-scene interaction features and geometry features. Together, they form a scene feature $\bm{F}_t \in \mathbb{R}^{256}$ at each time step $t$.

We first detail our BPS representation. Given an initialized motion sequence $\bm{X}_0, \bm{\hat{X}}_1, \bm{\hat{X}}_2, ..., \bm{\hat{X}}_T$, we use SMPL-X model~\cite{pavlakos2019expressive} to generate the corresponding human meshes $\bm{M}_0, \bm{M}_1, ..., \bm{M}_T$ where each $\bm{M}_{t} \in \mathbb{R}^{K \times 3}$ has $K$  downsampled vertices (we use $K=699$). To encode the human-scene interaction features, we compute $d(\bm{M}_t, \bm{S}) \in \mathbb{R}^{K \times 3}$, the difference between human vertices and their nearest neighbor points of the scene. As for the geometry scene features, we first define a cylinder with a radius $r=0.6$ m and a height $h=2$ m. The cylinder is centered at $(p_t^x, p_t^y, \frac{h}{2})$, where $p_t^x, p_t^y$ denote the horizontal position of the root at time step $t$. We sample $1024$ points from the volume of the cylinder at time $t$: $\bm{B}_t = [\bm{b}_1, \bm{b}_2, ..., \bm{b}_{1024}]^{T}$. We compute the geometry features of BPS representation as $d(\bm{B}_t, \bm{S}) \in \mathbb{R}^{1024 \times 3}$. We concatenate $d(\bm{M}_t, \bm{S})$ and $d(\bm{B}_t, \bm{S})$, and feed it to an MLP to get a lower dimensional vector as our final scene encoding $\bm{F}_t \in \mathbb{R}^{256}$ at time step $t$. 

To extract the scene features using PointNet++, we first pre-train a PointNet++ model for the semantic segmentation task on the S3DIS dataset~\cite{armeni20163d}. We input the scene point cloud $\bm{S}$ to the pre-trained model to get features for each point $\bm{F}^s \in \mathbb{R}^{N \times 128}$. For an arbitrary point $\bm{e}$ in 3D space, the feature $f(\bm{e}, \bm{F}^s) \in \mathbb{R}^{128}$ of $\bm{e}$ is computed by taking the inverse distance weighted interpolation, 
$f(\bm{e}, \bm{F}^s) =  \sum_{i=1}^{n_e}w_i\bm{F}^s(\bm{p}_{i})/\sum_{i=1}^{n_e}w_i$, 
where $w_i=1/||\bm{p}_i-\bm{e}||_2$, $n_e$ represents the number of nearest neighbors for $\bm{e}$ (we use three), and $\bm{p}_i$ represents the $i$th nearest neighbor point of $\bm{e}$. For the interaction features, we compute the feature at time step $t$ as $\bm{F}^h_{t} = [f(\bm{M}_{t}^{0}, \bm{F}^s), f(\bm{M}_{t}^{1}, \bm{F}^s), ..., f(\bm{M}_{t}^{K}, \bm{F}^s)]^T$ for each of the vertices $\bm{M}_t^k$ on the human mesh to obtain $\bm{F}^h_{t} \in \mathbb{R}^{K \times 128}$. For the geometry features, similarly, we use the nearest neighbor points of $\bm{B}_t$ denoted as $\bm{N}_t$ to compute the features as $\bm{F}^c_{t} = [f(\bm{N}_{t}^{0}, \bm{F}^s), f(\bm{N}_{t}^{1}, \bm{F}^s), ..., f(\bm{N}_{t}^{1024}, \bm{F}^s)]^T$ and obtain $\bm{F}^c_{t} \in \mathbb{R}^{1024 \times 128}$. Finally, we compute the mean of $\bm{F}^h_{t}$ and $\bm{F}^c_{t}$, and concatenate them to get a 256-dimensional vector as our final scene encoding at time step $t$, $\bm{F}_t \in \mathbb{R}^{256}$.

\paragraph{Model Architecture.}
We adopt a Transformer-based model architecture~\cite{vaswani2017attention} (Figure \ref{fig:overview}) to estimate human motions from the initialized motion $\bm{X}_0, \bm{\hat{X}}_1, ..., \bm{\hat{X}}_T$ and scene features $\bm{F}_0, \bm{F}_1, ..., \bm{F}_T$. Our model consists of four self-attention blocks, each containing a multi-headed attention layer followed by a position-wise feed-forward layer.  

\paragraph{Training Loss.}
Our training loss consists of four terms, including $L_1$ loss for root translation, global joint positions, local joint rotation, and joint positions computed using forward kinematics. The loss for each time step is defined as
\begin{align*}
    L = w&_{\text{trans}}\|\bm{p}_t - \bm{\hat{p}}_t\|_1 + w_{\text{joint}}\|\bm{J}_t - \bm{\hat{J}}_t\|_1 \\
    & + w_{\text{rot}}\|\bm{R}_t - \bm{\hat{R}}_t\|_1 + w_{\text{FK}}\|\bm{J}_t - \text{FK}(\hat{R}_t, \bm{\hat{p}}_t)\|_1,
\end{align*}
where $w_{\text{trans}}$, $w_{\text{joint}}$, $w_{\text{rot}}$, and $w_{\text{FK}}$ represent the loss weights for each term.  Note that our loss function does not leverage the scene information for supervision in order to have a fair comparison with our baselines. However, in practice these supervisions can and should be leveraged.

\section{Evaluation}
\label{sec:experiments}

We evaluate our model by measuring the generated motions using a set of standard metrics, by comparing against two baselines, and by inspecting the results visually. Please also view the supplemental video for more qualitative evaluation.

\begin{figure*}
    \centering
    \includegraphics[width=\linewidth]{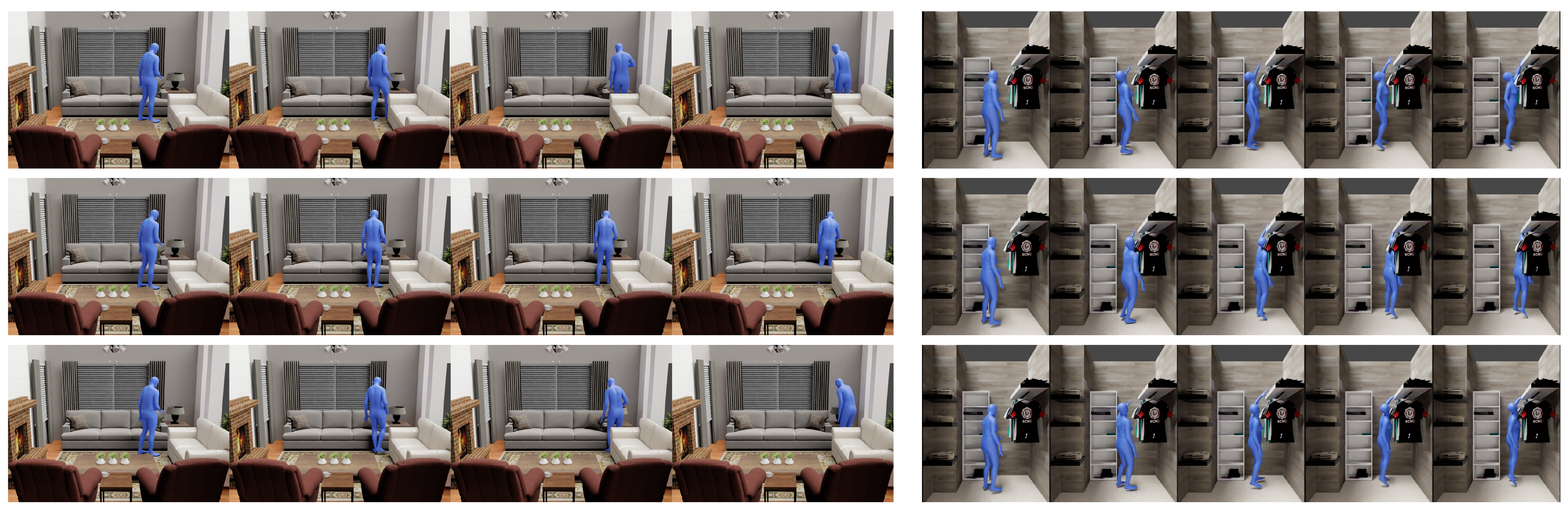}
    \caption{Comparison of the ground truth motion (top row) and the output generated by our model both with (bottom row) and without (middle row) scene conditioning for two different scenes (we show BPS scene condition only). We see that introducing scene information reduces collisions. For the full-resolution image and comparison with GOAL and PointNet scene representation please see the Supplementary Material.}
    \label{fig:results}
\end{figure*}

\begin{table*}[t]
    \centering
    \setlength{\tabcolsep}{5.5pt}
\begin{tabular}{llcccccccc}
\toprule
\multicolumn{2}{c}{} & \multicolumn{2}{c}{Task} & \multicolumn{2}{c}{Motion quality} & \multicolumn{4}{c}{Similarity to dataset} \\
\cmidrule(lr){3-4}\cmidrule(lr){5-6}\cmidrule(lr){7-10}
\multicolumn{1}{p{0.6cm}}{\centering Test split} & Model & \multicolumn{1}{p{1cm}}{\centering Success rate (\%)} & \multicolumn{1}{p{1cm}}{\centering Dist. to goal (cm)} & \multicolumn{1}{p{1.7cm}}{\centering Cumulative collision depth (cm)} & \multicolumn{1}{p{1cm}}{\centering Foot sliding (\%)} & \multicolumn{1}{p{1cm}}{\centering MVPE (cm)} & \multicolumn{1}{p{1cm}}{\centering MJPE (cm)} & \multicolumn{1}{p{1.5cm}}{\centering Root trans. error (cm)} & \multicolumn{1}{p{1cm}}{\centering Joint orn. error} \\
\midrule
\multirow{5}{*}{\rotatebox[origin=c]{90}{Random}} & Ground truth & 100.00 & 0.00 & 11.02 & 2.14 & 0.00 & 0.00 & 0.00 & 0.00 \\
 & NO-SCENE & 74.56 & \textbf{7.94} & 15.87 & 18.89 & 11.36 & 8.18 & 13.45 & 0.22 \\
 & GOAL & 12.14 & 44.48 & 20.89 & \textbf{12.78} & 17.78 & 21.48 & 19.99 & 1.97 \\
 & Ours (BPS) & 74.12 & 8.23 & 12.09 & 17.76 & \textbf{10.45} & \textbf{7.62} & \textbf{11.96} & 0.22 \\
 & Ours (PN2) & \textbf{79.20} & 8.22 & \textbf{11.63} & 18.83 & 11.09 & 8.05 & 12.60 & 0.22  \\
 \bottomrule
\end{tabular}
    \caption{Evaluation metrics on the random split (``PN2'' refers to PointNet++ scene encoding).}
    \label{tab:random_split_experiments}
\end{table*}

\begin{table*}[t]
    \centering
    \setlength{\tabcolsep}{5.5pt}
\begin{tabular}{llcccccccc}
\toprule
\multicolumn{2}{c}{} & \multicolumn{2}{c}{Task} & \multicolumn{2}{c}{Motion quality} & \multicolumn{4}{c}{Similarity to dataset} \\
\cmidrule(lr){3-4}\cmidrule(lr){5-6}\cmidrule(lr){7-10}
\multicolumn{1}{p{0.6cm}}{\centering Test split} & Model & \multicolumn{1}{p{1cm}}{\centering Success rate (\%)} & \multicolumn{1}{p{1cm}}{\centering Dist. to goal (cm)} & \multicolumn{1}{p{1.7cm}}{\centering Cumulative collision depth (cm)} & \multicolumn{1}{p{1cm}}{\centering Foot sliding (\%)} & \multicolumn{1}{p{1cm}}{\centering MVPE (cm)} & \multicolumn{1}{p{1cm}}{\centering MJPE (cm)} & \multicolumn{1}{p{1.5cm}}{\centering Root trans. error (cm)} & \multicolumn{1}{p{1cm}}{\centering Joint orn. error} \\
\midrule
\multirow{5}{*}{\rotatebox[origin=c]{90}{Task}} & Ground truth & 100.00 & 0.00 & 8.17 & 2.11 & 0.00 & 0.00 & 0.00 & 0.00 \\
 & NO-SCENE & 64.92 & 9.08 & 10.18 & 18.95 & 11.61 & 8.38 & 14.11 & 0.23 \\
 & GOAL & 8.43 & 47.60 & 18.05 & \textbf{11.24} & 17.27 & 20.94 & 20.40 & 1.95 \\
 & Ours (BPS) & \textbf{76.08} & \textbf{8.30} & \textbf{7.04} & 20.66 & 11.47 & 8.34 & \textbf{13.13} & 0.23 \\
 & Ours (PN2) & 70.84 & 9.19 & 9.22 & 18.97 & \textbf{11.32} & \textbf{8.14} & 13.47 & 0.23 \\
 \bottomrule
\end{tabular}
    \caption{Evaluation metrics on the task split (``PN2'' refers to PointNet++ scene encoding).}
    \label{tab:task_split_experiments}
\end{table*}

\subsection{Setup}

We train our model on CIRCLE, use two different criteria to split the data, and compare against two baselines. To keep all sequences under $240$ frames (the size of our model input), we downsample every sequence to $20$ FPS. We train our model for $1800$ epochs using AdamW~\cite{loshchilov2017decoupled} with weight decay $0.01$ and an initial learning rate of $0.0001$ that is multiplied by $0.3$ every $1000$ epochs. 

\paragraph{Dataset Split.} We evaluate on two separate splits:
\begin{itemize}
    \item \textbf{Random}: We randomly split the reaching sequences into $2565$ for training and $453$ for testing.
    \item \textbf{Task}: We hold out specific tasks (start/goal region pairs) for testing, resulting in $2578$ sequences for training and $440$ for testing ($109$ and $19$ tasks, respectively).
\end{itemize}

\paragraph{Baselines.}
We compare our approach with two baselines:
\begin{itemize}
    \item \textbf{GOAL:} The MNet described in GOAL \cite{taheri2021goal}, an MLP architecture that, given a start and a goal pose\footnote{We provide the ground truth goal pose instead of training GOAL's GNet module to make the baseline stronger. The original architecture conditions the output on the BPS representation of an object that is to be grasped. We modify it to condition on the BPS representation of the environment.}, autoregressively predicts the next pose in the sequence. 
    \item \textbf{NO-SCENE:} Our architecture without the scene encoding $\bm{F}_0, \cdots \bm{F}_T$. Since our loss function does not depend on the scene information, our model does not have any privileged supervision over this baseline.
\end{itemize}
All baseline methods are trained on the same data.

\subsection{Evaluation Metrics}
\label{sec:experiments:eval_metrics}

We use the following metrics to evaluate our method.

\paragraph{Task Completion.} Task success is determined by a threshold on euclidean distance between the generated and ground truth position of the right-hand wrist at the last frame in the sequence. An instance of a task is considered successful if this distance is lower than $10$ cm.

\paragraph{Collision Avoidance.} We check each frame for collisions between the human mesh and the scene and compute the sum of interpenetration depths as our collision metric. To empirically correct for MoSh fitting inaccuracies, we filter out collisions with a depth lower than $2$ cm.

\paragraph{Foot Sliding.} Our chosen metric is percentage of frames in a sequence with sliding. Heuristically, a frame is considered to have sliding if the velocity of the vertex with the lowest height is greater than $1$ cm per frame. Critically, this heuristic applies to sequences where the human is kneeling or lying down rather than standing.

\paragraph{Similarity to Dataset.} To measure the similarity to the ground truth motion, we calculate the mean vertex error, mean joint position error, mean root translation error, and mean joint orientation error as the Frobenius norm of the rotation matrix representation of the root's orientation.

\begin{figure}
    \centering
    \includegraphics[width=\linewidth]{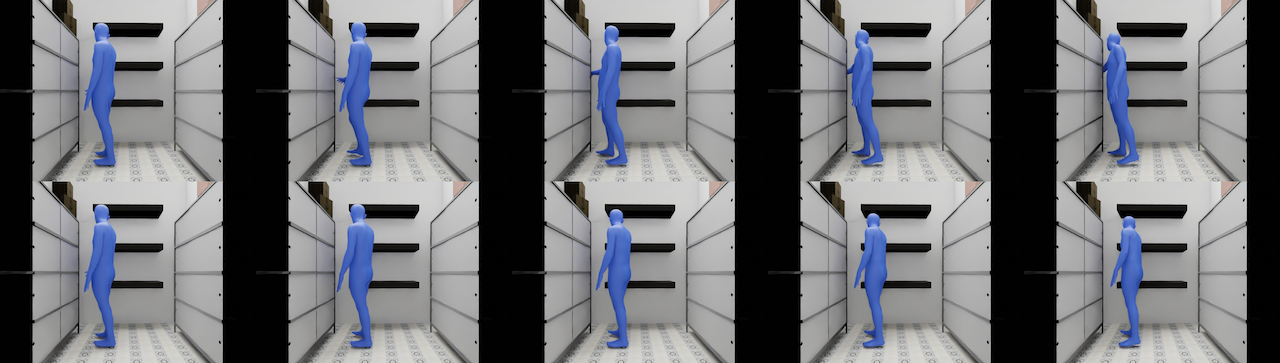}
    \caption{GOAL baseline test sequence (bottom) and corresponding ground truth motion (top). The model moves the character very little (especially the feet), resulting in poor task and similarity metrics, despite low prevalence of foot sliding artifacts.}
    \label{fig:goal_baseline}
\end{figure}

\subsection{Results}
We calculate each metric from \ref{sec:experiments:eval_metrics} for every sequence and report its average value over all sequences in \cref{tab:random_split_experiments} and \cref{tab:task_split_experiments}. 
\paragraph{Task Completion.}
Our model clearly outperforms the GOAL baseline in the task completion metric (\cref{tab:random_split_experiments}). We attribute this to the fact that GOAL has to predict the sequence autoregressively, and so all the steps need to contribute towards reaching the goal position. 
In contrast, our method can, by design, accomplish long-term, scene conditioned planning via full sequence prediction.
Note that, despite the absence of a goal loss term in our training procedure, our models are still able to score high in the task metrics with joint position losses. Our model also moderately outperforms the NO-SCENE baseline. 

\paragraph{Collision Avoidance.}
Our model clearly outperforms NO-SCENE in collision avoidance (\cref{tab:random_split_experiments}). Figure \ref{fig:results} demonstrates the visual comparison of the differences. Since our model is not supervised by any measure of collision avoidance in the loss function, this gain can be attributed to scene state observations. Despite having access to the same scene information, GOAL sequences have deeper collisions.

\paragraph{Foot Sliding.}
GOAL produces significantly fewer foot sliding artifacts than our approach. However, visualizing the GOAL predicted sequences  (\cref{fig:goal_baseline}) reveals that the model often outputs a nearly static pose for all frames, explaining this deceptive contrast. Our approach with scene conditioning generally produces more foot sliding artifacts than NO-SCENE. This is possibly attributable to the greater overall adaptation of source data to fit scene constraints.

\paragraph{Similarity to Dataset.}
Although similarity to ground truth is not the primary goal, it is a loose metric that indicates the overall quality of the human motion. Our model and NO-SCENE perform similarly, while GOAL is trailing behind. We encourage readers to view the supplemental videos for additional evaluation of the motion quality.

\paragraph{Unseen Tasks}
Table \ref{tab:task_split_experiments} illustrates generalization to unseen tasks. As expected, overall performance for all models have degraded on this split. However, we continue to observe our approach notably outperforming GOAL in task completion. Additionally, our approach with BPS continues to outperform NO-SCENE on collision depth.

\section{Conclusion}

We introduce an efficient new method of capturing contextual motion data within a realistic cluttered environment in virtual reality, leverage this system to generate a novel contextual motion dataset, CIRCLE, and use this dataset to train models that generate scene-aware human motion.

While our first application of CIRCLE demonstrates its potential, generalized contextual motion generation remains a challenging problem. Our evaluation indicates that more work will be necessary from the research community to produce better adapted motion with fewer artifacts.

In addition to improving on our results, we would like to understand how the data collected with our system compares with human motion in real physical environments. We are also interested in studying how generalization improves with more data. Fortunately, our streamlined capture pipeline will enable us to continue expanding upon the initial scale and diversity of CIRCLE.

Looking to the future, we are interested in exploring more interactive contextual tasks. For example, expanding our system to capture physical and virtual object trajectories in addition to human motion. This next step toward interactive data collection has the potential to fill a critical need in Embodied AI, fueling efforts to understand how humans interact with objects in their everyday lives and pass those skills to artificial, embodied agents.

\paragraph{Acknowledgments.} This work is in part supported by Meta AI, NSF CCRI \#2120095, ONR MURI N00014-22-1-2740, and the Stanford Institute for Human-Centered AI.

\clearpage

%%%%%%%%% REFERENCES
{\small
\bibliographystyle{ieee_fullname}
\bibliography{egbib}
}

\clearpage

\appendix

\begin{figure*}
    \centering
    \includegraphics[width=\linewidth]{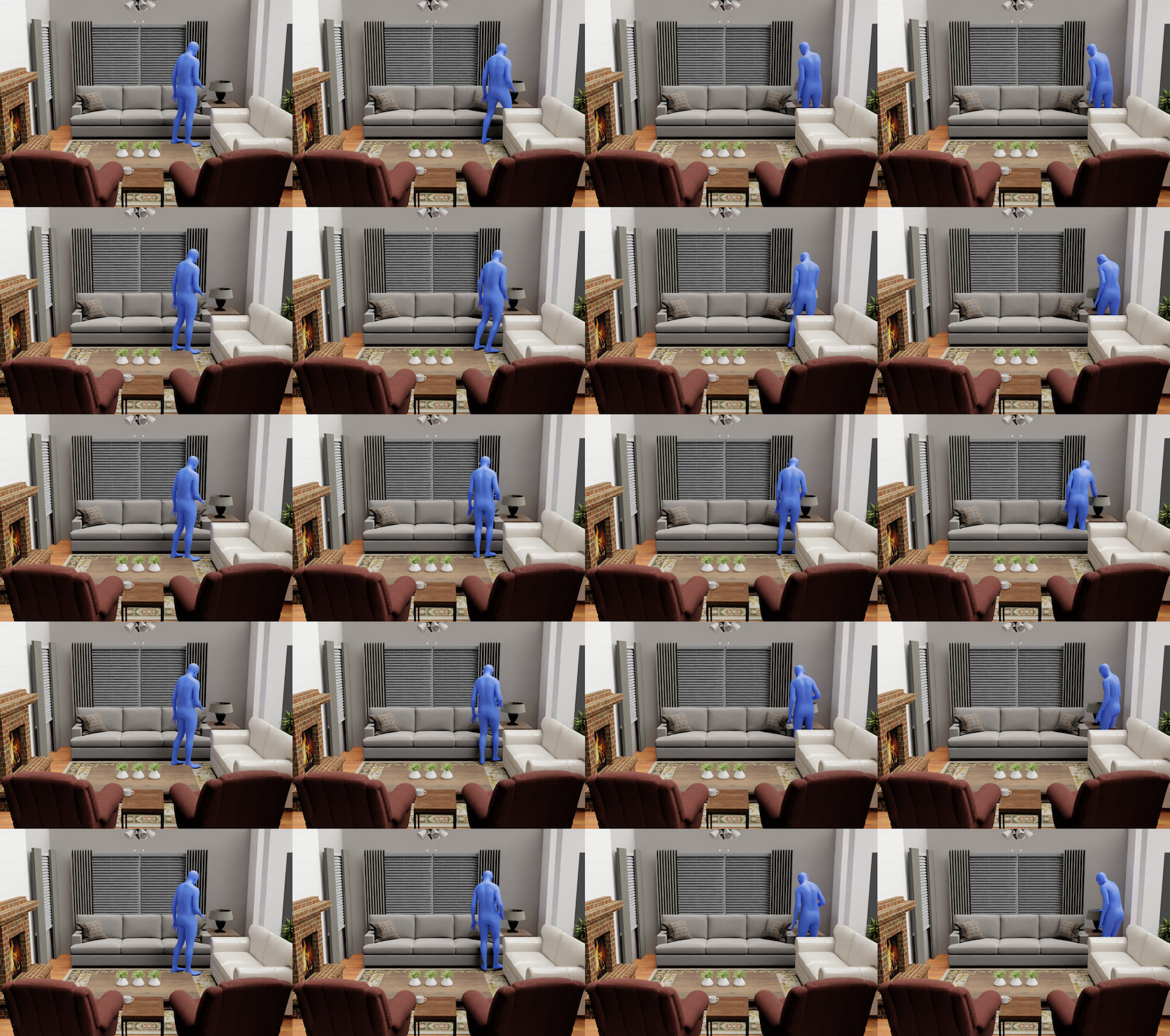}
    \caption{Comparison of the i) ground truth motion (top row), and outputs generated by ii) GOAL (second row), iii) our model without scene information (third row), iv) scene information using pointnet (fourth row), v) scene information using BPS (fifth row). We see that the sequence generated by GOAL fails to achieve the objective, and that introducing scene information reduces collisions.}
    \label{fig:results_sup_living}
\end{figure*}

\begin{figure*}
    \centering
    \includegraphics[width=\linewidth]{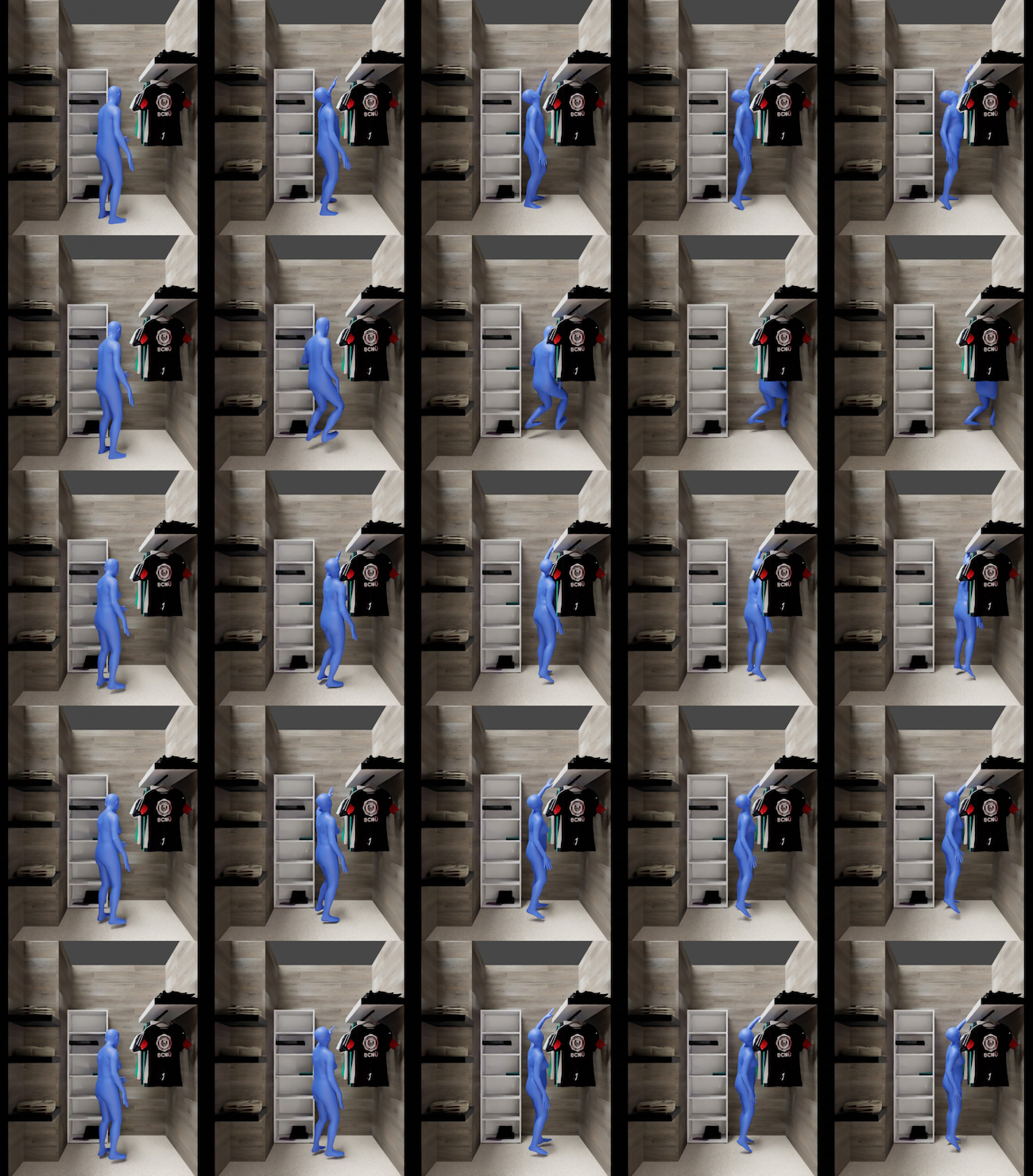}
    \caption{Comparison of the i) ground truth motion (top row), and outputs generated by ii) GOAL (second row), iii) our model without scene information (third row), iv) scene information using pointnet (fourth row), v) scene information using BPS (fifth row). We see that the sequence generated by GOAL fails to achieve the objective, and that introducing scene information reduces collisions.}
    \label{fig:results_sup_bed}
\end{figure*}
\section{Supplemental}

\subsection{Skeleton/headset alignment}

We use the headset's built-in calibration tool to align its forward direction with the forward direction of the motion capture volume. We also calibrate the mocap system and the headset such that they measure the same floor height. This reduces the alignment problem to finding a planar offset $\vec{o}$ (we do not modify the skeleton's height) that is used to align the livestreamed skeleton to the actor in VR. To achieve this, we assume that the midpoint of the eyes (measured by the headset),

$$
\vec{r}_e = \frac{\vec{r}_{\mbox{left eye}} + \vec{r}_{\mbox{right eye}}}{2},
$$
lies along a line that has the direction of the head bone's local forward vector and contains the midpoint of the head bone's top face ($\vec{f}_{h}$ and $\vec{r}_{ht}$, both respectively measured by the mocap system). We can write this as a linear system with $3$ constraints and $3$ unknowns,

$$
\vec{r}_e = \vec{r}_{ht} + \lambda \vec{f}_{h} + \vec{o},
$$
where

$$
\vec{o} = \begin{bmatrix}
    o_x \\ o_y \\ 0
\end{bmatrix},
$$
that we solve during calibration.

\section{High resolution result images}

For the complete versions of the images in \cref{fig:results}, see \cref{fig:results_sup_living} and \cref{fig:results_sup_bed}, respectively.

\end{document}